# Bond-Centered Molecular Fingerprint Derivatives: A BBBP Dataset Study

Guillaume Godin[1]

[1]Osmo Labs PBC New York, USA

Corresponding authors: guillaume@osmo.ai

**Abstract**

- Bond Centered FingerPrint (BCFP) are a complementary, bond-centric alternative to Extended-Connectivity Fingerprints (ECFP) (1). We introduce a static BCFP that mirrors the bond-convolution used by directed message-passing GNNs like ChemProp (2), and evaluate it with a fast rapid Random Forest model (3) on Brain-Blood Barrier Penetration (BBBP) classification task. Across stratified cross-validation, concatenating ECFP with BCFP consistently improves AUROC and AUPRC over either descriptor alone, as confirmed by Turkey HSD multiple-comparison analysis. Among radii, $r = 1$ performs best; $r = 2$ does not yield statistically separable gains under the same test. We further propose BCFP-Sort&Slice, a simple feature-combination scheme that preserves the out-of-vocabulary (OOV) count information native to ECFP count vectors while enabling compact unhashed concatenation of BCFP variants. We also outperform the MGTP(4) prediction on our BBBP evaluation, using such composite new features bond and atom features. These results show that lightweight, bond-centered descriptors can complement atom-centered circular fingerprints and provide strong, fast baselines for BBBP prediction.

**Scientific Contribution**

- **Bond-Centered Fingerprints (BCFP).** We introduce a bond-centric extension to ECFP and show that concatenating BCFP + ECFP yields higher BBBP classification performance than either alone, with statistically significant gains in AUROC/AUPRC (Tukey HSD on CV folds).

- **OOV-Sort&Slice.** We generalize the Sort&Slice scheme to preserve out-of-vocabulary (OOV) counts inherent to ECFP.

- **Radius insight.** We find radius 1 BCFP provides the best trade-off on BBBP, while radius 2 does not achieve a statistically separable improvement under the same evaluation.

- **Benchmarking.** Our BCFP descriptors contribute to surpass the MGTP baseline on BBBP using a lightweight XGBoost pipeline, highlighting the strength of simple, fast models with well-designed features.

- **Comparability of GNN layers.** We clarify that Chemprop's directed bond-centric message passing is not layer-wise comparable to node-centric MPNNs; we provide a matched-budget evaluation protocol and discuss why "same number of layers" can be misleading for cross-architecture comparisons.

# Introduction

A strong and fast baseline in molecular property prediction is a Random Forest (RF) trained on ECFP4/ECFP6 descriptors. In practice, the count-based variant of ECFP generally outperforms the binary variant, especially for classification. Recent deep-learning approaches can match or exceed these baselines, including pretrained transformer–CNN models (5) and graph neural networks such as ChemProp or AttentiveFP(6).

Chemprop's key architectural choice is directed, bond-centered message passing, in contrast to the more common atom-centered formulations used by many MPNNs. Because much of the remaining architecture is comparable across message-passing GNNs, this raises a focused question: what concrete advantage does the bond-centered formulation confer over atom-centered approaches?

To isolate this representational factor, we introduce a static Bond-Centered Fingerprint (BCFP) that mirrors Chemprop's bond-centric view, and we compare it directly against ECFP using a lightweight Random Forest or XGBoost pipeline on the Blood–Brain Barrier Penetration (BBBP) classification task. To our knowledge, this is the first study to propose BCFP and analyze its complementarity with ECFP (7). Our results indicate that concatenating atom- and bond-centered fingerprints yields efficient and effective models for BBBP prediction, clarifying why bond-centric message passing often appears among top-k performers while offering a simple, fast alternative to full neural architectures.

# Data

We use the original BBBP dataset from Martins et al. (8) containing 2050 molecules. After removing the invalid SMILES and duplicates, we obtain 1957 unique molecules. We also evaluate a second larger BBBP dataset coming from a recent study (4) with 4094 unique molecules, produced via a stringent cleanup from B3DB study (9,10).

Inspired by Walters (11), rather than a 5x5 CV, we perform 29 seeds independent random splits to obtain paired statistics for feature comparisons (the goal is comparison, not a single best model). For each seed in ensemble {0,28}, we create a 80/20 train - test split using scikit-learn (12).

We implement BCFP in both python and cpp with RDkit (13) and compute count-based vectors for the following configurations at radii r, r in {0,1,2,3}.

- ECFP(r)
- BCFP(r)
- EFCP(r) || BCFP (r) (concatenation)
- EFCP(r)|| BCFP(r-1) (hybrid) for r>0 else EFCP(0) || BCFP (0)

This yields 4 fingerprints setups x 4 radius x 29 seeds.

Unless noted, we used Random Forest as a fast baseline. We also compute an exact 5x5 CV on the 1957 molecule set. We also report the 4094 molecule set using 10x10 CV for random forest and XGBoost (14). We extend Sort&Slice to demonstrate the generalization of our method.

## Pure random replicates

After training the 464 Random-Forest we evaluated AUROC and F1 metrics (figure 1 and 2). We can get the same conclusion based on the two metrics. Radius r = 1 is generally better to others, while r = 2 is very similar to r = 1. Concatenation of "EFCP || BCFP" is slightly better than the fingerprints separately.

Because "radius" mirrors the number of message-passing steps, these results align with known depth effects (over-smoothing/squashing) and help explain why Chemprop (directed, bond-centered) often excels at shallow depth. Practically, a 2-layer Chemprop is not directly comparable to a 2-layer atom-centered MPNN (15); a fairer match is Chemprop (L) ≈ atom-centered MPNN (L+1).

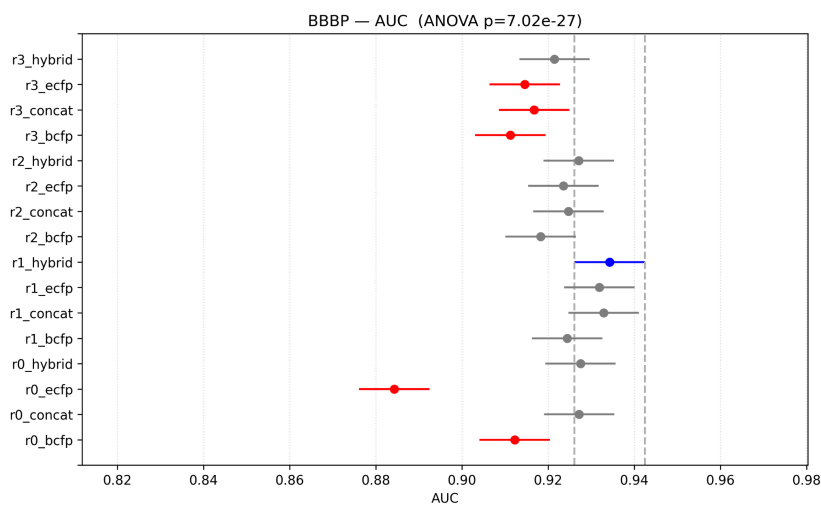

Figure 1 : AUROC score of Random forest on 29 random seed splits of 20% on 1957 BBBP molecules

The largest gap appears at radius 0: as expected, BCFP(0) is already more informative than ECFP(0) across all metrics, supporting our hypothesis that bond-centric encoding > atom-centric encoding at very small neighborhoods. Strikingly, at radius 0 the concatenated and hybrid descriptors (ECFP(r) || BCFP(r) and ECFP(r) || BCFP(r−1)) perform close to the best radii, whereas BCFP(0) and ECFP(0) alone are the weakest features. This indicates strong complementarity: atom- and bond-centered views recover missing local context when combined, even with minimal radius.

Beyond r = 1, performance degrades. We attribute this primarily to hash collisions: increasing the radius expands the number of distinct substructure keys while the fingerprint dimensionality remains fixed, raising collision rates and reducing signal-to-noise.

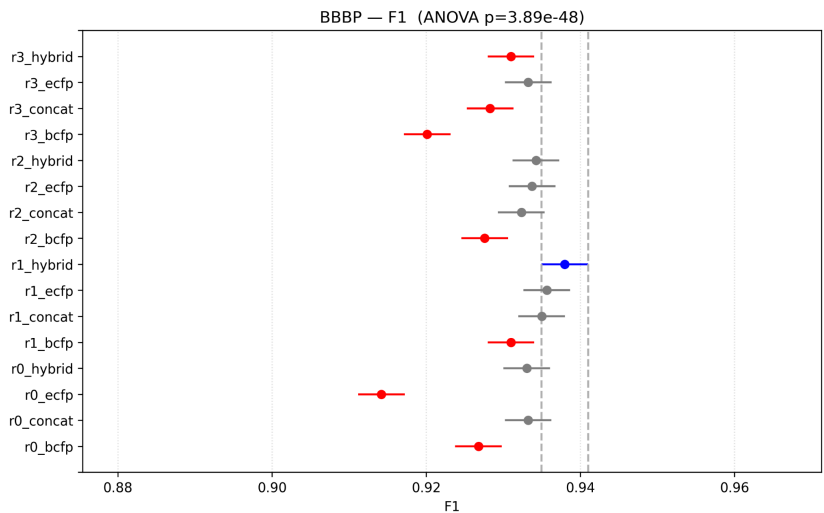

Figure 2 score of Random forest on 29 random seed splits of 20% on 1957 BBBP molecules

# True 5 x 5 CV

We ran a stratified 5x5 cross validation on the 1957 molecule set. Results closely mirror the 29-seed random split experiment (Table 1). The radius-1 hybrid descriptor (ECFP(1) || BCFP(0)) yields the best AUROC, though margins are small. Despite minimal context, concat (ECFP(r) || BCFP(r)) and hybrid (ECFP(r) || BCFP(r−1)) are already competitive at r=0, reinforcing the complementarity of atom- and bond-centered views.

Overall, the agreement between 5×5 CV and the 29-seed analysis strengthens our conclusion that combining ECFP + BCFP provides a small but consistent improvement for BBBP classification.

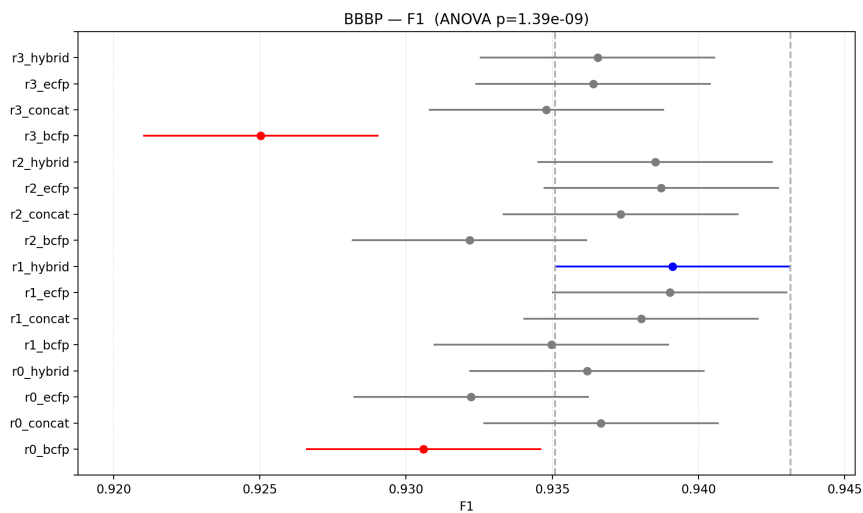

Figure 3: F1 score of Random forest on Stratigirf Kflod 5 x 5CV on 1957 BBBP molecules at 4 radius 0,1,2,3

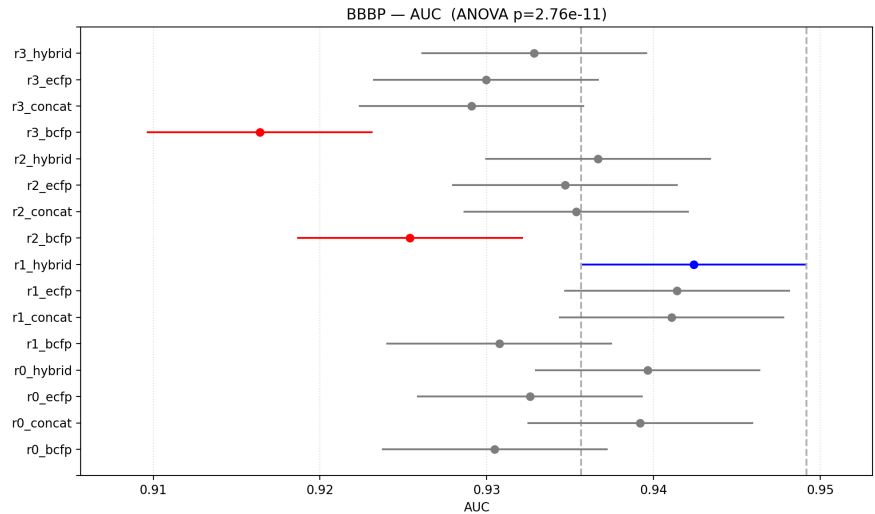

Figure 4 : AUROC score of Random forest on StratifiedKflod 5 x 5CV on 1957 BBBP molecules at 4 radius 0,1,2,3

# Extend to Sort&Slice

To align with the BBBP 4094 molecule set, we ran repeated StratifiedKfold 10x10. Because the original seeds are unavailable, we used our own 10 seeds, while preserving stratification within each of the 10 folds per repeat. The goal is to feature benchmarking against original MGTP results, without hyperparameter chasing(16).

We extended both ECFP and BCFP with Sort&Slice (_ss) using "count" instead of "binary" (consistent with released code, even if the paper is not 100% clear on best method), producing new descriptors:

- ECFP_ss(r)
- BCFP ss(r)
- Concat: ECFP_ss(r) ∥ BCFP_ss(r)
- Hybrid: ECFP_ss(r) ∥ BCFP_ss(r−1) (for r>0 else ECFP_ss(r) ∥ BCFP_ss(r)

Winner is r = 1 concat (ECFPss(1) || BCFPss(1)) descriptor is equivalent to slightly better than the MGTP fingerprint random forest performance (AUPRC: 0.9016+/-0.0183 & AUC: 0.8495+/-0.0266) as shown in figures 5 and 6.

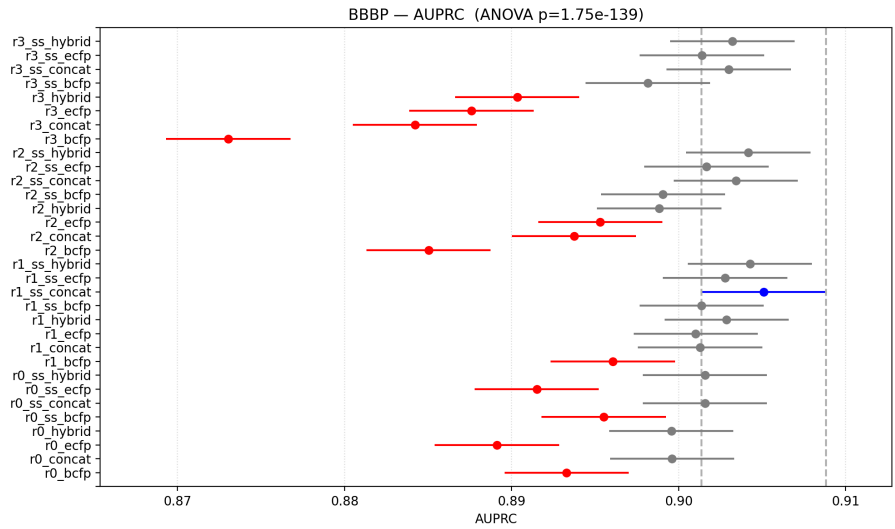

Figure 5 : AUPRC score of Random forest on StratifiedKflod 10 x 10 CV on 4094 BBBP molecules at 4 radius 0,1,2,3 including adapted version of sort&slice count ECPF and BCFP.

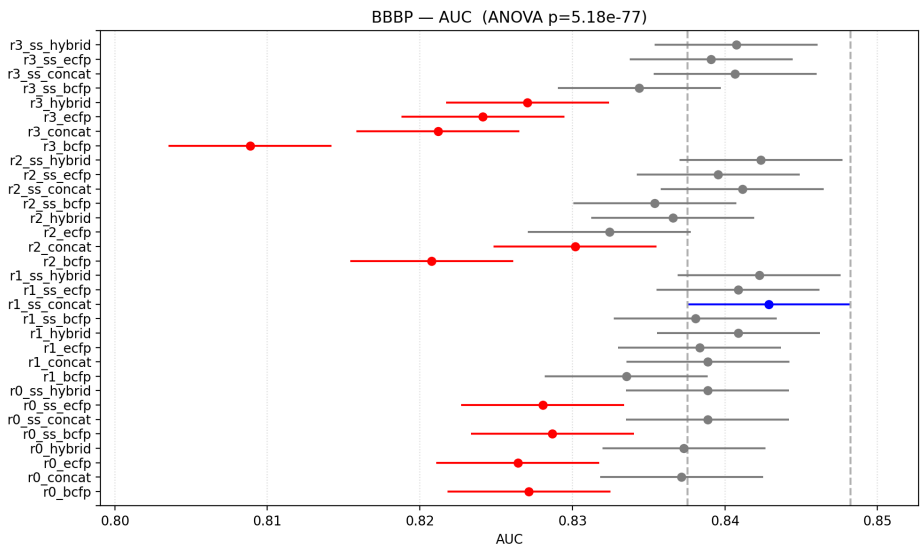

Figure 6: AUROC score of Random forest on StratifiedKflod 10 x 10 CV on 4094 BBBP molecules at 4 radius 0,1,2,3 including adapted version of sort&slice count ECPF and BCFP.

We repeated the 10x10 stratified CV experiment on the 4094 molecule BBBP dataset using XGBoost, to compare against the MGTP XGBoost baseline (AUPRC = 0.9115+/-0.0155, AUC = 0.8638+/-0.0242 for XGBoost).

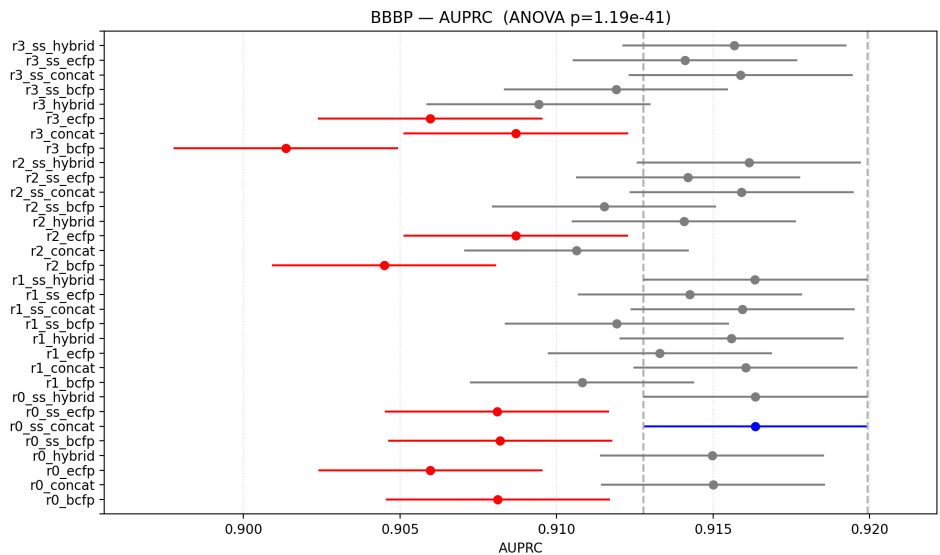

Figure 7 : AUPRC score of XGBoost on StratifiedKflod 10 x 10 CV on 4094 BBBP molecules at 4 radius 0,1,2,3 including adapted version of sort&slice count ECPF and BCFP.

Extending ECFP and BCFP to count-based Sort&Slide ("ss_") yields clear gains. Both "ECPF_ss(r) || BCPF_ss(r)" and "ECPF_ss(r) || BCPF_ss(r-1)" outperform their classical (hashed) counterparts, and the margin widens as radius increases. For both metrics, we get better results than the benchmark MGTP best model reported. The worst performance occurs at very low radii (insufficient context) and high radii for hashed fingerprints (collision burden). Sort&Slice mitigates the latter, maintaining strong performance at r = 1 (and preventing the steep drop seen with hashed features as r grows)

Increasing radius expands the set of substructure keys while fixed-length hashed vectors raise collision rates, degrading signal (an effect analogous to over-smoothing/blurring in deeper GNN message passing). Counted Sort&Slice preserves discriminative detail and curbs collision-driven loss, explaining its advantage with XGBoost see figures 7 and 8.

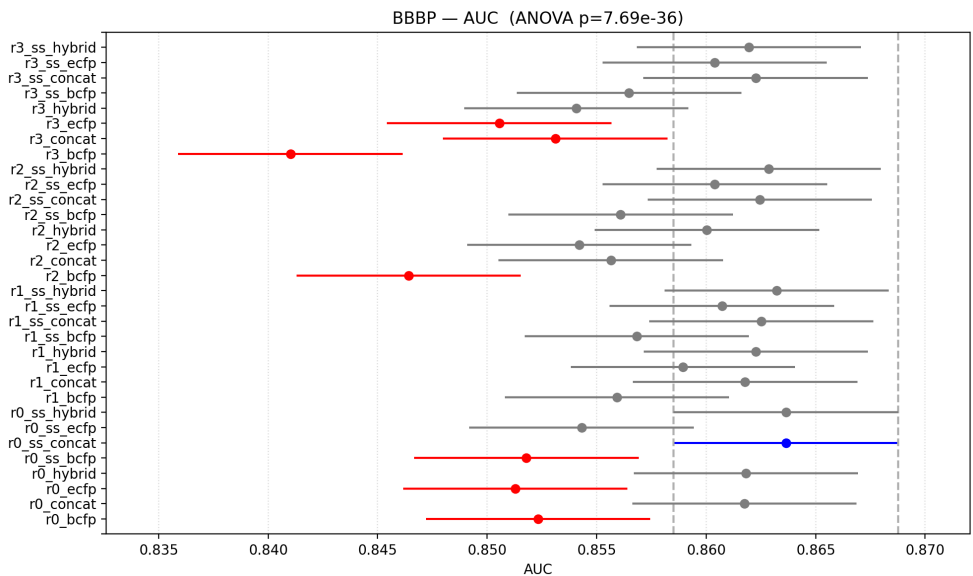

Figure 8 : AUROC score of XGBoost on StratifiedKflod 10 x 10 CV on 4094 BBBP molecules at 4 radius 0,1,2,3 including adapted version of sort&slice count ECPF and BCFP.

# Fine tune Sort&Slice method

We augment Sort&Slice with a single out-of-vocabulary (OOV) bucket: an extra coordinate that accumulates the counts of all keys not retained in the top-K slice (including keys that would otherwise be "unseen/unknown"). This preserves information that hashed counts normally keep (but plain Sort&Slice discards), thereby restoring complementarity between ECFP and BCFP at larger radii.

Using Random Forest models, we can see stabilization of OOV_Sort&Slide descriptors performances (_ss) in figures 9 & 10 compared to figures 5 & 6.

Adding a single OOV coordinate is a minimal change that recovers dropped signal, stabilizes performance, making OOV_Sort&Slide less sensitive to radius extension.

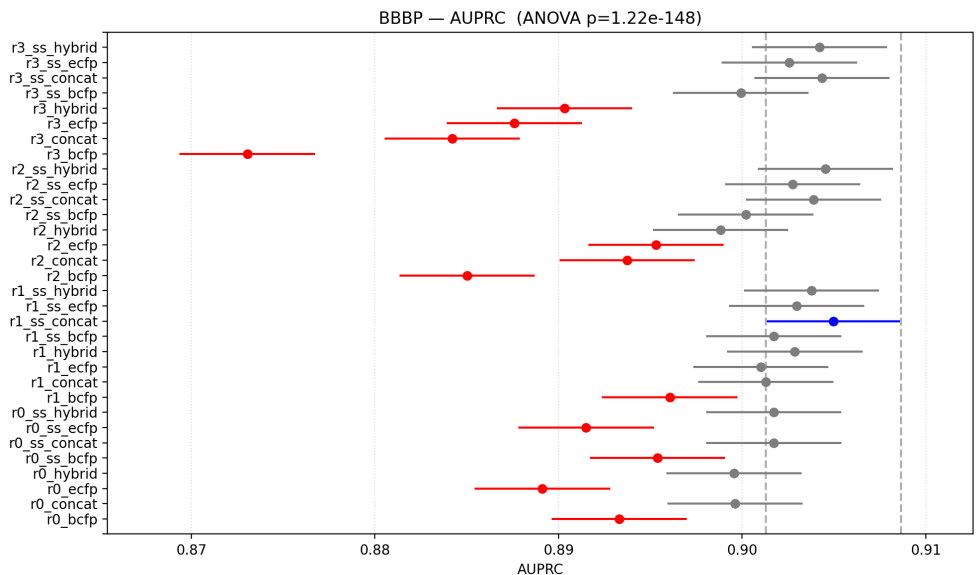

Figure 9 : AUPRC score of Random Forest on StratifiedKflod 10 x 10 CV on 4094 BBBP molecules at 4 radius 0,1,2,3 including fine tune sort&slice count ECPF and BCFP.

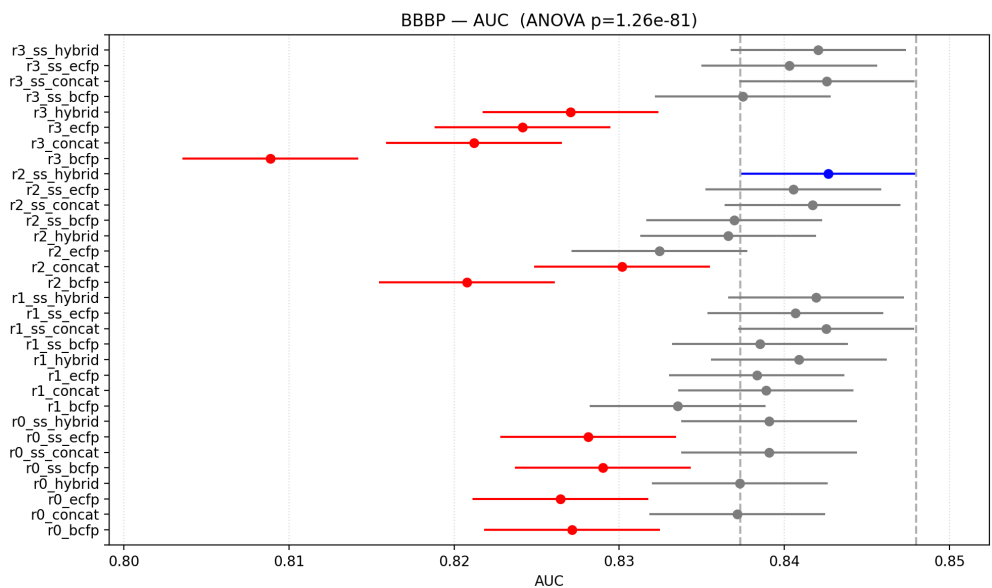

Figure 10 : AUROC score of Random Forest on StratifiedKflod 10 x 10 CV on 4094 BBBP molecules at 4 radius 0,1,2,3 including fine tune sort&slice count ECPF and BCFP.

# Discussion

A bond intrinsically encodes both participating atoms *and* their connectivity, so a bond-centered view can be more expressive than an atom-centered one at the same neighborhood size. Empirically we observe a radius-shift: BCFP(r) ≈ ECFP(r+1), i.e., bond-centered features reach a given performance at one smaller radius. Consistent with this, the hybrid descriptor BCFP(r−1)

|| ECFP(r) tends to edge out the same-radius concat BCFP(r) || ECFP(r); this is visible (albeit modestly) in Tukey HSD analyses of the F1 plots. These observations support the initial hypothesis that bond-centric encodings carry richer local information than atom-centric encodings.

Because "radius" mirrors message-passing depth, layer-wise comparisons across architectures can be misleading. A fairer match is approximately Chemprop(L) (directed, bond-centered) vs. atom-centered MPNN(L+1), rather than L vs. L.

Our Sort&Slice implementation uses per-molecule key enumeration (as in the original method) rather than global enumeration, which can introduce split dependence. To reduce this, we add a single OOV bucket that accumulates counts of keys dropped by the slice or unseen in a split. This change stabilizes performance, especially at larger radii where classical hashed fingerprints suffer from hash collisions; with OOV-Sort&Slice, the r=3 degradation is markedly reduced and results become more uniform across r=1–3.

# Conclusion

Combining ECFP + BCFP yields consistently better BBBP prediction than either alone. The trend is visible across metrics; AUROC/AUPRC generally separate configurations best, with F1 corroborating the same ordering.
Shallow neighborhoods work best: r=1 is the sweet spot. Prefer hybrid (BCFP(r−1) || ECFP(r)) or concat (BCFP(r) || ECFP(r)) over single-view descriptors.
Results help explain why bond-centered, directed message passing (e.g., Chemprop) often performs strongly at small depth, while also suggesting that mixed atom+bond designs could be advantageous. Notably, shallow models like AttentiveFP often use only two convolution layers by default (17)
The proposed OOV-Sort&Slice reduces radius sensitivity and mitigates collision-driven blur at larger r, acting as a lightweight regularizer that improves robustness and test-set applicability.

Using our new advanced bond and atom fingerprint allows us to surpass the MGTP performance on both random forest and XGBoost.

# Funding

The author was funded by Osmo Labs PBC for BCFP evaluation.

# Competing Interests and Consent for publication

# Acknowledgement

The author wants to thank Brian Kelley and Gregory Landrum for the discussion on BCFP implementation during the hackathon day at last RDkit UGM 2025 at Prague.

# SUPPLEMENTARY

| Fingerprint | AUC | F1 | Radius |
|---|---|---|---|
| ECFP | 0.915±0.020 | 0.933±0.008 | 3 |
| BCFP | 0.911±0.019 | 0.920±0.005 | 3 |
| Concat | 0.917±0.019 | 0.928±0.005 | 3 |
| Hybrid | 0.922±0.018 | 0.931±0.006 | 3 |
| ECFP | 0.924±0.019 | 0.934±0.006 | 2 |
| BCFP | 0.918±0.017 | 0.928±0.005 | 2 |
| Concat | 0.925±0.017 | 0.932±0.006 | 2 |
| Hybrid | 0.927±0.017 | 0.934±0.006 | 2 |
| ECFP | 0.932±0.015 | 0.936±0.007 | 1 |
| BCFP | 0.924±0.017 | 0.931±0.006 | 1 |
| Concat | 0.933±0.016 | 0.935±0.006 | 1 |
| Hybrid | 0.934±0.016 | 0.938±0.008 | 1 |
| ECFP | 0.884±0.019 | 0.914±0.009 | 0 |
| BCFP | 0.912±0.019 | 0.927±0.008 | 0 |
| Concat | 0.927±0.017 | 0.933±0.006 | 0 |
| Hybrid | 0.928±0.017 | 0.933±0.007 | 0 |

Table 1: F1 & AUROC scores of Random forest on 29 random seed splits of 20% on 1957 BBBP molecules at 4 radius 0,1,2,3 (in blue best/ red worse)